# CONVERSION OF BRAILLE TO TEXT IN ENGLISH, HINDI AND TAMIL LANGUAGES

S.Padmavathi[1], Manojna K.S.S[2], Sphoorthy Reddy .S[3] and Meenakshy.D[4]

Amrita School of Engineering, Amrita Vishwa Vidyapeetham, Coimbatore, India

[1]s_padmavathi@cb.amrita.edu, [2]manojna.kapala@gmail.com,
[3]sphoorthy.surakanti@gmail.com, [4]meenakshymenon14@gmail.com,

## ABSTRACT

.The Braille system has been used by the visually impaired for reading and writing. Due to limited availability of the Braille text books an efficient usage of the books becomes a necessity. This paper proposes a method to convert a scanned Braille document to text which can be read out to many through the computer. The Braille documents are pre processed to enhance the dots and reduce the noise. The Braille cells are segmented and the dots from each cell is extracted and converted in to a number sequence. These are mapped to the appropriate alphabets of the language. The converted text is spoken out through a speech synthesizer. The paper also provides a mechanism to type the Braille characters through the number pad of the keyboard. The typed Braille character is mapped to the alphabet and spoken out. The Braille cell has a standard representation but the mapping differs for each language. In this paper mapping of English, Hindi and Tamil are considered.

## 1. INTRODUCTION

Visually impaired people are an integral part of the society. However, their disabilities have made them to have less access to computers and Internet than the people with clear vision. Over time Braille system has been used by them for written communication. Braille is a system of writing that uses patterns of raised dots to inscribe characters on paper. This allows visually impaired people to read and write using touch instead of vision. It is the way for blind people to participate in a literate culture. First developed in the nineteenth century, Braille has become the pre-eminent tactile alphabet. Its characters are six- dot cells, two wide by three tall as shown in figure 1. Each dot may exist or may not exist giving two possibilities for each dot cell. Any of the six dots may or may not be raised; giving 64 possible characters. In English it includes 26 English alphabets, punctuations, numbers etc. Figure 2 shows the Braille representation of English alphabets. Braille representation for numerals is shown in figure 3.

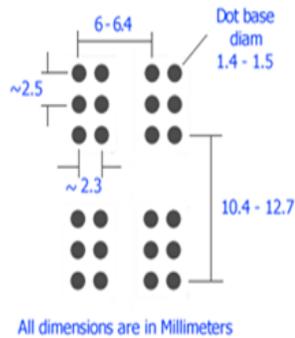

Fig1:Braille cell dimensions          Fig 2.Braille characters of Alphabet

Fig 3. Braille characters for numerals          Fig 4. Contraction

Although Braille cells are used world-wide, the meaning of each cell depend on the language that they are being used to depict. In English Braille there are three levels of encoding: Grade 1, a letter-by-letter transcription used for basic literacy; Grade 2, an addition of abbreviations and contractions; and Grade 3, contains over 300 abbreviations and contractions that reduce the amount of Braille codes needed to represent written text. Some of the contracted words are represented in figure 4 in its Braille format.

Braille can be seen as the world's first binary encoding scheme for representing the characters of a writing system. However, very limited numbers of Braille books are available for usage. Printing of Braille books is a time consuming process. The requirement of special printers and software add to their limited availability. Scanned and text converted documents can be used in the meantime to serve the needs of the blind.

This paper mainly focuses on conversion of a Braille document into its corresponding alphabets of three main languages namely English, Tamil and Hindi using various concepts of image processing. The presence of dots in the Braille cells has to be identified to recognize the characters. The edge detection when applied on the scanned document will not produce the dots, hence the approximate intensity range of the dots are identified from the histogram. The image is treated through a sequence of enhancement steps which increases the contrast between the dots and the background. The edge detectors are then applied and the text area is cropped excluding the borders through projection profile method. The document is then segmented into Braille cells using standard Braille measurements and projection profiles. The presence of dots in each cell is identified using a Threshold and converted to Binary sequence which is then mapped to the corresponding language alphabet. This paper also proposes a Number keypad which could be used for typing the analogous Braille alphabet using six numbers i.e. (7,4,1,8,5,2) corresponding to the six dot cells.

Section 2 of this paper discusses about few commercial systems available for converting Braille to text or vice-versa. Section 3 explains about the conversion of Braille to text; section 4 covers the outcomes of the proposed method. Section 5 concludes the paper.

## 2. LITERATURE SURVEY

A Braille translator is a software program that translates a script into Braille cells, and sends it to a Braille embosser, which produces a hard copy in Braille script of the original text. Basically only the script is transformed, not the language.

One of the general purpose translators is text to Braille converter. Few other commercial translators are also available such as win Braille, supernova, cipher Braille translator and Braille master.

### 2.1. Text to Braille converter:

Displays Braille as the user types characters. This convertor is OS independent and language used is java. It concentrates on conversion from English to Braille

### 2.2. Win Braille:

As referred in [2] Win Braille can be used without prior Braille knowledge. It includes standard Windows image control and the unique feature to convert images to tactile graphic format on-line.

### 2.3. Braille Master:

As referred in [5] the Braille Master package comes with both Windows and DOS versions. A large print facility suitable for partially sighted persons is also included in this package.

### 2.4. Cipher Braille Translator:

As referred in [2] Cipher is a text to Braille program that converts text documents into a format suitable for producing Braille documents, through the use of a Braille printer. The user can edit, save, use style templates and enable translation rules.

### 2.5. Supernova:

As referred in [2] Supernova is a window-based magnifier, screen reader and a Braille system that supports the conversion of text to speech, Braille displays and note-takers. Braille can be converted to text using number keypad and image processing techniques which are feasible for common people.[8] refers to a paper on Braille word segmentation and transformation of Mandarin Braille to Chinese character. [9] discuss the main concepts related to OBR systems;

list the work of different researchers with respect to the main areas of an OBR system, such as pre-processing, dot extraction, and classification. [10] describes an Arabic Braille bi-directional and bi-lingual translation/editor system that does not need expensive equipments. [11] focuses on developing a system to recognize an image of embossed Arabic Braille and then convert it to text. [12] presents a new Braille converter service that is a sample implementation of scalable service for preserving digital content. [13] proposes a software solution prototype to optically recognise single sided embossed Braille documents using a simple image processing algorithm and probabilistic neural network. [14] describes an approach to a prototype system that uses a commercially available flat-bed scanner to acquire a grey-scale image of a Braille document from which the characters are recognised and encoded for production or further processing. [15] introduces a new OBR system which designed for recognizing a scanned Arabic Braille document and converting it into a computerized textual form that could be utilized by converting it into voice using other applications, or it could be stored for later use. [16] presents an automatic system to recognize the Braille pages and convert the Braille documents into English/Chinese text for editing.[17] describes a new technique for recognizing Braille cells in Arabic single side Braille document. [18] describes the character recognition process from printed documents containing Hindi and Telugu text. [19] involves a keyboard which is a device made of logical switches and uses Braille system technique for sensing the characters[20] develope a system that converts, within acceptable consrtraints , (Braille image) to a computer readable form. [21] describes the Sparsha toolset. [22] presents a system for a design and implementation of Optical Arabic Braille Recognition(OBR) with voice and text conversion. [24] provides a detailed description of a method for converting Braille as it is stored as characters in a computer into print. [25] describes a new system that recognizes Braille characters from image taken by a high speed camera to Chinese character and at the same time automatically mark the Braille paper.

## 3. BRAILLE CONVERSION

For converting the Braille document to text, the input is taken in two different formats. In the first method the Braille character is accepted as a sequence of numbers typed through the keypad and in the second method a scanned Braille document is taken as input. The Braille character is extracted in each case and matched with the corresponding alphabet with help of a pre built Trie structure.

### 3.1. Keypad to type Braille document

The six dot cell representation of Braille character could be numbered from 1 to 6 starting from top left to bottom right in the order left to right and top to bottom. The numbers 7,4,1,8,5,2 of keypad are mapped to the dots 1,2,3,4,5,6 respectively as shown in fig 5.

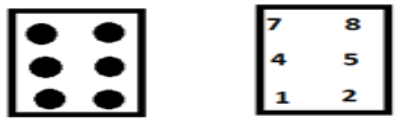

Fig 5. Mapping of dots to numbers

With the number pad the number sequences of the Braille characters are typed and used for further conversion. The number sequences of the English Braille alphabets are listed in Table 1 and that of contractions are shown in Table 2.

Table 1: Mapping of Alphabets

| Character | Represe-ntation | Character | Represe-ntation |
|---|---|---|---|
| A | 7 | N | 7851 |
| B | 74 | O | 751 |
| C | 78 | P | 7841 |
| D | 785 | Q | 78451 |
| E | 75 | R | 7451 |
| F | 784 | S | 841 |
| G | 7845 | T | 8451 |
| H | 745 | U | 712 |
| I | 84 | V | 7412 |
| J | 845 | W | 8452 |
| K | 71 | X | 7812 |
| L | 741 | Y | 78512 |
| M | 781 | Z | 7512 |
| : | 45 | ; | 41 |
| " | 412 | ! | 451 |
| , | 4 | . | 452 |

Table 2: Mapping of contracted words

| Contracted word | Corresponding representation |
|---|---|
| but | 74 |
| can | 78 |
| from | 748 |
| for | 741852 |
| of | 74152 |
| the | 4182 |
| with | 41852 |
| ed | 7482 |
| er | 74852 |

In this implementation, 0 is used as delimiter for an alphabet, 3 for word and 6 for the end of sentence. A trie is created for the number sequence from top left to the bottom right with numbers of rows below become the decedents of numbers of rows above. The alphabet of a corresponding number sequence is stored as a leaf. This trie is pre created and used for matching and recognition of Braille alphabets.

Matching of the number sequence is done as the numbers are typed and the corresponding alphabet is displayed when a delimiter is encountered. A voice corresponding to the alphabet is delivered as a feedback to the user. A beep is sound in case of an error. This enables the person to rectify the alphabet immediately.

### 3.2. Conversion of scanned Braille document

In this method the Braille document is scanned and taken as input, which by a sequence of steps is converted to appropriate text. The scanned document has to be enhanced to identify the dots clearly. The dots are extracted using horizontal and vertical profiling. The Braille cells are identified and converted to binary sequence. The binary sequence is then mapped to the corresponding alphabets or contracted words. These are stored in a text file and given as input to the voice synthesizer. The basic block diagram is shown in figure 6.

Utmost care is taken to ensure that unwanted noise or redundant information is not introduced at the time of scanning. The scanned image is then converted to gray scale image.

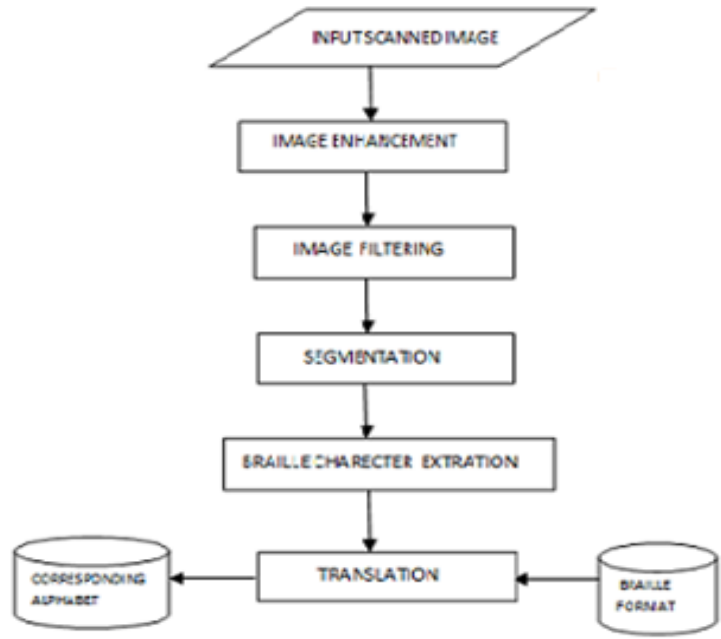

Fig 6.Block Diagram of Proposed Technique

### 3.2.1. Image enhancement

Due to scanning, the dots in the Braille document cannot be distinguished clearly from the background. Hence various pre processing techniques are applied on the scanned image in order to enhance the dots and to suppress the noise. The dots appear as a darker shade of the back ground color and hence these intensity ranges are identified from the Histogram and enhanced in order to identify the dots. Piece wise enhancement techniques such as contrast stretching, intensity stretching were used for enhancing the dots. These techniques could be represented as $S=T(r)$, where S is the grey level after modification T is the enhancement function used and r is grey level before enhancement.

Contrast stretching is the process that expands the range of the intensity levels in an image as shown in figure 7.This is used to enhance the slightly dark dots from the background. The limits over which image intensity values will be extended are decided from the histogram of the input image.

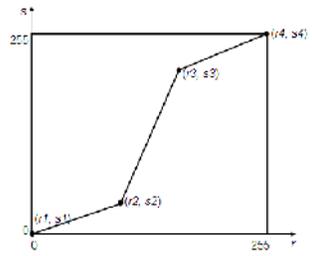 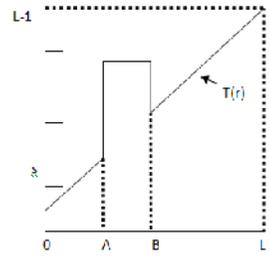

Fig7: Contrast Stretching    Fig8: Intensity Adjustment

Another level enhancement is done to the dots using Intensity adjustment. This is an image enhancement technique that maps the intensity values of an image to a new range as shown in figure 8. This transformation highlights intensity ranges [A, B] and preserves all other intensity levels.

### 3.2.2. Image filtering

To remove the unwanted noisy dots present in the scanned documents, the image is smoothened using Gaussian filter and then subjected to morphological opening using a disk shaped structure element B as given in the equation 1

$$A \circ B = (A \ominus B) \oplus B \qquad (1)$$

Where $\ominus$ denotes erosion.

An edge detected binary image is obtained using a Prewitt filter. The Prewitt operator uses two 3×3 kernels which are convolved with the image A, to calculate approximations of two derivatives - one for horizontal changes, $G_x$, and one for vertical, $G_y$.

$$G_x = \begin{bmatrix} -1 & 0 & +1 \\ -1 & 0 & +1 \\ -1 & 0 & +1 \end{bmatrix} * A \quad \text{and} \quad G_y = \begin{bmatrix} +1 & +1 & +1 \\ 0 & 0 & 0 \\ -1 & -1 & -1 \end{bmatrix} * A \qquad (2)$$

Where * denotes convolution.

The resulting gradient approximations can be combined to give the gradient magnitude, using Eq (3). When magnitude is greater than the threshold T, it is identified as an edge.

$$G = \left(G_x^2 + G_y^2\right)^{0.5} \qquad (3)$$

The edges mostly correspond to the dots of the Braille cells. The border of scanned document and the stapler pin information if any present in the document are removed through image cropping.

### 3.2.3. Segmenting the Braille Cells

In order to simplify the process of Braille character extraction, the image is first segmented into lines and then into Braille cells. Each cell is further partitioned into binary dot patterns. These are achieved through Projection profiles and standard Braille measurements. Horizontal profiling is performed on edge detected image and zero profile indicates the absence of dots and hence the line as shown in Fig 9. Among many such lines, the first line from the top that is closer to the dots is taken as reference. The standard vertical distance between two Braille cells is used to draw the remaining lines where the X projection is zero. This procedure is repeated till the end of the document.

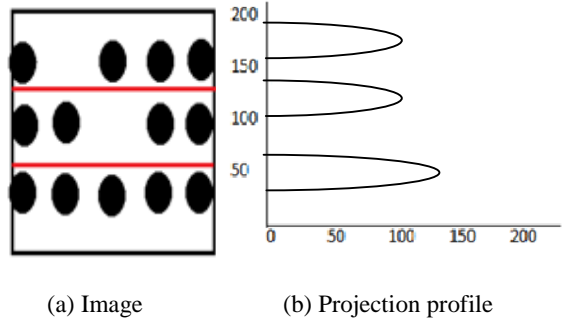

(a) Image          (b) Projection profile

Fig9: Horizontal projection profiling

After extracting horizontal lines of the Braille cell, a vertical profiling is performed. Zero profiles indicate the vertical lines as shown in Fig 10. Among many such lines the leftmost one that is closer to the dots is taken as reference. The standard horizontal distance between two Braille cells is used to draw the remaining lines where the Y projection is zero. This procedure is repeated till the end of the document. This segments the edge image in to Braille cells. Each segmented cell is divided into 3 x 2 grids using the standard Braille distance between two dots in a cell.

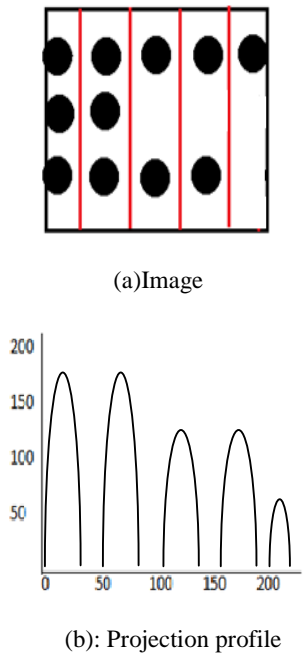

(a) Image

(b): Projection profile

Fig10: Vertical Projection Profile

International Journal of Computer Science, Engineering and Applications (IJCSEA) Vol.3, No.3, June 2013

### 3.2.4. Extraction of Text from pattern vector

A Binary pattern vector for each Braille cell is generated. A vector has a length of 6 each correspond to a dot in the Braille cell. The presence of dot is identified after counting the number of white pixels in each grid of a cell and checking whether it satisfies the threshold criterion. '1' indicates that dot is present and '0' indicates that dot is absent in that particular position. This string of bits for the sequence of Braille alphabets is written into a file. A sequence of 6 bits are read from the file and converted to the number sequence and subsequently into the alphabet using the trie structure as discussed in 3.1. If the six bits of the string are 0's, it generates a space. These alphabets are stored in a text file for further processing. Natural Reader [22] is called for reading the converted English text. For Tamil and Hindi, the sequences of 6 bits are taken and the corresponding Unicodes are generated using its pre built mapping table. These Unicodes are stored in the file. The obtained file is converted to the corresponding Tamil and Hindi text. eSpeak[25] is used to read the converted Tamil and Hindi text.

## 4. EXPERIMENTAL ANALYSIS

The data set includes 20 Braille documents among which 10 are Grade 2 English documents, 5 sheets are Hindi and 5 Tamil sheets. Grade 2 includes contractions which are explained in section 1. Fig11 shows the scanned Braille document. Fig12 shows the edge detected image without applying any enhancement or noise removal techniques. Fig13 shows the edge detected image after applying image enhancement and cropping as explained in section 3.2.

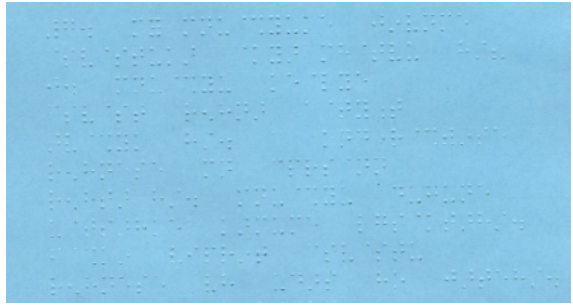

Fig11. Scanned Braille document

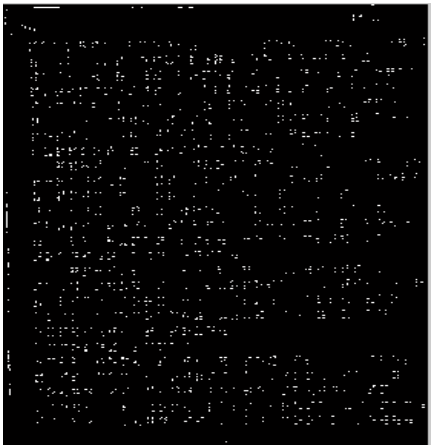   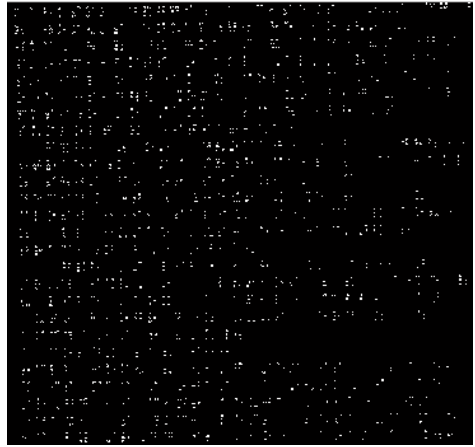

   Fig12. Edge detected image with noise        Fig13. Edge detected image without noise

International Journal of Computer Science, Engineering and Applications (IJCSEA) Vol.3, No.3, June 2013

Fig14 shows the image with each line of the document separated by red lines after performing the horizontal profiling. The sample corresponds to Grade2 Braille document. Fig15 shows the image after horizontal and vertical profiling with boxes drawn for each Braille cell. Fig16 focuses on a particular cell after drawing a grid for extracting the dots.

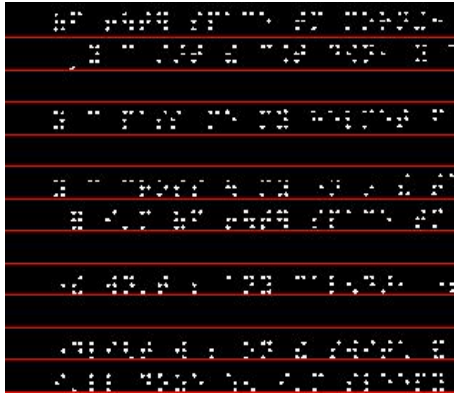
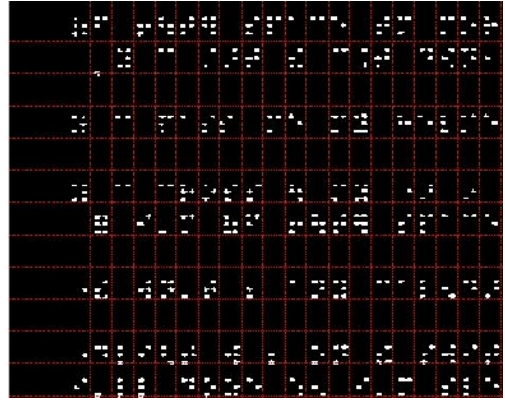

Fig14. Image after horizontal profiling    Fig15. Image after horizontal and vertical profiling

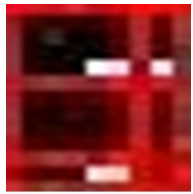

Fig16. Extraction of each dot in a cell

After the extraction of the dots, a binary sequence is generated as shown in Fig17 where value '1' shows dot is present and '0' shows dot is absent in a cell. The English alphabet sequence mapped from the binary sequence is shown in Fig18. The mapped alphabets for Tamil and Hindi for two other documents are shown in Fig19, Fig 20 respectively. The errors occurring due to wrong punching of Braille characters are indicated in blue colour box. The errors occurring while combining the Unicodes are indicated in red colour box. When a short notation of a word is used, those are not expanded by the system. For example 'tm' is a short notation used for 'tomorrow'. These are indicated in green box. In Hindi document, appropriate spacing was not punched in the input Braille document and hence the words appear consecutively.

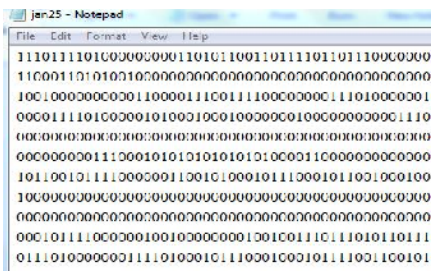
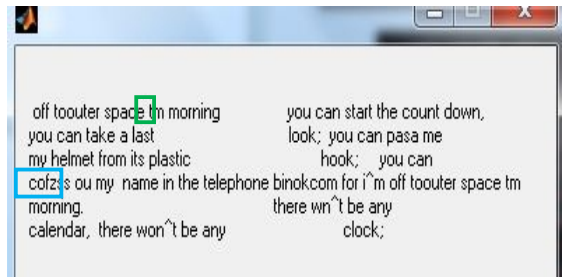

Fig17. Binary sequence for English    Fig18. Final mapped text for English

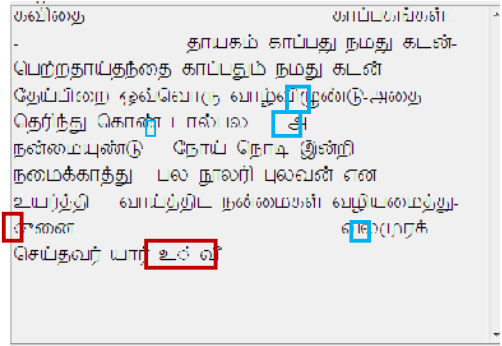 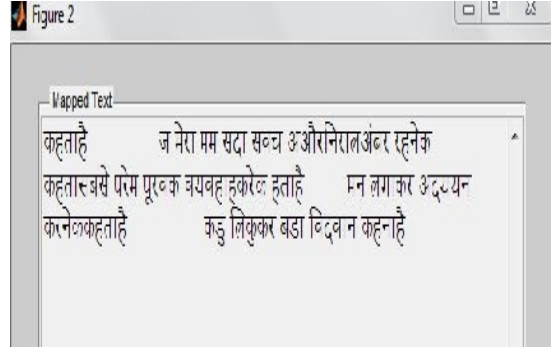

Fig19. Final mapped text for Tamil   Fig20. Final mapped text for Hindi

To evaluate the performance of the system, each Braille document is decoded manually and compared with the system results. The effect of various enhancement techniques on the system performance is tabulated for English document in Table3. The table shows the percentage of accuracy of the words correctly identified after mapping. In the table CS represents contrast stretching, IS represents intensity stretching and MO represents morphological operations. The sequence of image enhancement techniques when applied on the image also influences the accuracy of the words mapped. The accuracy for few sequences is shown in column 2 through 4. The accuracy is found to be high for the sequence CS, IS, and MO. The accuracy drops drastically when the order of enhancement techniques is changed. These are illustrated in the table. The number of words correctly mapped and the percentage of accuracy for the English(E#), Tamil(T#) and Hindi(H#) documents are shown in Table 4. In this table, TW represents the total number of words present in the Braille document and C represents the number of words correctly mapped by the system, final column specifies the percentage of accuracy in mapping. Confusion occurs when two different symbols have same Braille representation. For example in English, ',' and 'ea' ; 'be' and ';' in Tamil letter ரு and symbol ':' has same braille representation. Accuracy drops when such confusion occurs. In Hindi the percentage of accuracy is dropped because of the manual mistakes done while punching the Braille document.

Table 3: Analysis of enhancement Techniques.

|        | CS  | CS,IS | CS,IS,MO | MO,CS,IS |
|--------|-----|-------|----------|----------|
| Doc 1  | 75% | 86%   | 97%      | 32%      |
| Doc 2  | 72% | 88%   | 97%      | 34%      |
| Doc 3  | 80% | 85%   | 98%      | 36%      |
| Doc 4  | 71% | 80%   | 97%      | 34%      |
| Doc 5  | 75% | 86%   | 93%      | 24%      |
| Doc 6  | 73% | 83%   | 97%      | 39%      |
| Doc 7  | 77% | 82%   | 96%      | 26%      |
| Doc 8  | 70% | 89%   | 95%      | 33%      |
| Doc 9  | 80% | 89%   | 96%      | 32%      |
| Doc 10 | 78% | 88%   | 91%      | 29%      |

Table 4: Accuracy of text conversion.

|      | TW  | C   | Accuracy( %) |
|------|-----|-----|--------------|
| E 1  | 343 | 341 | 99.4         |
| E 2  | 350 | 346 | 98.2         |
| E 3  | 425 | 425 | 100          |
| E4   | 327 | 320 | 97.8         |
| E 5  | 402 | 398 | 99.0         |
| E 6  | 250 | 247 | 98.8         |
| E 7  | 361 | 354 | 98.0         |
| E 8  | 393 | 392 | 99.7         |
| E 9  | 285 | 283 | 99.2         |
| E 10 | 324 | 324 | 100          |
| T 1  | 378 | 375 | 99.2         |
| T 2  | 405 | 399 | 98.5         |
| T 3  | 290 | 287 | 98.9         |
| T 4  | 318 | 318 | 100          |
| T 5  | 328 | 326 | 99.3         |
| H 1  | 345 | 341 | 98.8         |
| H 2  | 323 | 320 | 99.0         |
| H 3  | 298 | 297 | 99.6         |
| H 4  | 276 | 272 | 98.5         |
| H 5  | 354 | 351 | 99.1         |

## 5. LIMITATIONS and ADVANTAGES

Since the standard Braille dimensions are used for the segmentation of the Braille cells, the document has to be free from tilt and has to be aligned with the edge of the scanner. This poses a major limitation to the system. The presence of the unnecessary dots or noises whose size is comparable to that of the Braille dots during scanning is difficult to remove during pre processing and hence affects the accuracy of the converted text.

It involves very less intervention of the user and helps to serve the need of large number of people using a single document. It helps resource teachers in Inclusive Education, who do not know Braille. Simplifies making of copies of old Braille books for which only one copy is available as it saves the labour of preparing the same again. Since the availability of Braille document is also limited, scanning the document also help in preserving the existing documents.

## 6. CONCLUSION

This paper focuses on the conversion of scanned Braille documents to corresponding text in English language and Indian languages namely Hindi and Tamil. After identifying the start of the Braille text, the lines and subsequently the Braille cell are segmented. Grids are drawn based on the standard measurement of the Braille cells and the dots are extracted. Braille has a standard pattern of alphabets and only the mapping differs from language to language. Using appropriate mapping for each language the alphabets are identified and stored as text. These texts are read out by voice synthesizer. The extraction of the dots was affected when they were not confined to the standard measurement and due to the presence of noise during scanning. Mapping errors occurred when the Braille has similar representation for the alphabet and the punctuation. These are eliminated to some extent using simple rules governing the language. The mapping errors are predominant for Grade 2 English documents. The voice synthesizer used for speaking the Native languages had a poor pronunciation. The paper could be extended for Grade 3 English documents and the voice synthesizer for Hindi and Tamil could be customized.